\def\year{2021}\relax
\documentclass[letterpaper]{article} 
\usepackage{aaai20}  
\usepackage{times}  
\usepackage{helvet} 
\usepackage{courier}  
\usepackage[hyphens]{url}  
\usepackage{graphicx} 
\usepackage{booktabs}
\usepackage{amsfonts}
\usepackage{graphicx}  
\title{Primer AI's Systems for Acronym Identification and Disambiguation}
\author{Nicholas Egan, John Bohannon\\ 
Primer AI\\ 
San Francisco, CA\\
\{negan, bohannon\}@primer.ai 
}
 
\begin{document}

\maketitle

\begin{abstract}
The prevalence of ambiguous acronyms make scientific documents harder to understand for humans and machines alike, presenting a need for models that can automatically identify acronyms in text and disambiguate their meaning. We introduce new methods for acronym identification and disambiguation: our acronym identification model projects learned token embeddings onto tag predictions, and our acronym disambiguation model finds training examples with similar sentence embeddings as test examples. Both of our systems achieve significant performance gains over previously suggested methods, and perform competitively on the SDU@AAAI-21 shared task leaderboard. Our models were trained in part on new distantly-supervised datasets for these tasks which we call AuxAI and AuxAD. We also identified a duplication conflict issue in the SciAD dataset, and formed a deduplicated version of SciAD that we call SciAD-dedupe. We publicly released all three of these datasets, and hope that they help the community make further strides in scientific document understanding.
\end{abstract}

\section{Introduction}
Writers of scientific documents frequently utilize abbreviations as tools to make unwieldy technical terms less verbose. These abbreviations often take the form of acronyms or initialisms, which are abbreviations formed from the first letters of words in the term. We refer to the abbreviated form as the ``short form'' or ``acronym,'' and we refer to the full term as the ``long form'' or ``expansion.'' The widespread usage of these abbreviations makes writing more convenient for scientists, but poses a challenge to machines and non-expert humans attempting to read scientific documents. This has led to an accumulation of scientific jargon, and a need for AI tools to manage acronyms and their expansions.

\citeauthor{veyseh-et-al-2020-what} (\citeyear{veyseh-et-al-2020-what}) recently released two large datasets for acronym understanding in scientific documents: the first is for the acronym identification task (AI), and the second is for the acronym disambiguation task (AD). The goal of acronym identification is to extract short and long form acronyms within a sentence, and the goal of acronym disambiguation is to determine the expansion of a particular acronym given sentence context.

\paragraph*{Contributions} In this paper, we describe our systems for the AI and AD tasks, which improve upon the models proposed by \citeauthor{veyseh-et-al-2020-what} and perform competitively on the task leaderboard \cite{shared-task}. Our AI method projects learned token embeddings from a transformer-based language model onto tag predictions, and our AD method finds similar training examples for testing examples. We improved the performance of our systems through the development of distantly-supervised auxiliary datasets, which we are releasing to the public. Finally, we identified some issues with the SciAD dataset, and propose a remedy for these issues that we hope will make SciAD more useful as a tool for the NLP community. Our three datasets are publicly available on our GitHub data repository.\footnote{https://github.com/PrimerAI/sdu-data}

\section{Datasets}
\subsection{SciAI}
The SciAI dataset \cite{veyseh-et-al-2020-what} consists of 17,560 sentences annotated for acronym identification, where each sentence token is tagged for short form and long form acronym boundaries in BIO format. To construct this dataset, the authors assembled a corpus of 6,786 papers from arXiv, identified candidate sentences in these papers that likely contained acronyms, and hired Amazon Mechanical Turk workers to gold label the sentences. The candidate sentences were sentences containing consecutive (or near consecutive) word sequences for which the concatenation of the first few characters from these words could spell out another word in the document that consists of at least 50\% capital letters. When labeling, humans were instructed to find all short form acronyms in the sentence, even if the acronym's long form did not appear in the sentence.

\subsection{Auxiliary AI Data}
To build on the training data provided by SciAI, we took a distantly supervised approach to build a more noisy dataset we call AuxAI. We started by scraping abbreviations with their expansions from \url{Abbreviations.com} within the Academic \& Science, Computing, and Internet categories. We then searched for these terms on arXiv, finding paper abstracts for which the short and long form both appear. We were able to identify 6497 terms across 274,149 abstracts this way. After finding these abstracts, we searched our abstracts for any other abbreviations that appear with their expanded form, and labeled them as such. Additionally, since some acronyms like USA and DNA are common enough to stand on their own in short form without including their expanded forms, we compiled a list of ``universal acronyms'' which are acronyms in the SciAI training dataset that have no long form in their sentence for at least 2 sentences. We found 1807 such acronyms, which were all marked in our token labels.

\begin{table}[ht]
\centering
\begin{tabular}{@{}lrrr@{}}
\toprule
\textbf{Tag} & \textbf{SciAI Train} & \textbf{SciAI Dev} & \textbf{AuxAI} \\ \midrule
O & 85.12\% & 84.78\% & 88.01\% \\
B-short & 5.34\% & 5.69\% & 3.59\% \\
I-short & 0.31\% & 0.35\% & 0.01\% \\
B-long & 2.94\% & 2.92\% & 2.55\% \\
I-long & 6.29\% & 6.27\% & 5.84\% \\ \bottomrule
\end{tabular}
\caption{Distribution of BIO tag labels among the training and development partitions of the SciAI dataset, as well as the AuxAI dataset.}
\label{tab:auxai-dist}
\end{table}

Overall, we were able to produce a dataset of consisting of 313,914 sentences. Table \ref{tab:auxai-dist} shows that the tag distribution within this dataset skews slightly more towards the O tag, suggesting that our distantly supervised approach has imperfect recall. Another issue with this dataset is that it contains around 3800 different short form acronyms, while the SciAI training set contains around 6500 despite being smaller, suggesting that AuxAI captures less data diversity. During training, we experimented with subsampling AuxAI data in such a way that the ratio between unique terms and training examples matched that of SciAI.

\subsection{SciAD}
The SciAD dataset \cite{veyseh-et-al-2020-what} consists of 62,441 examples annotated for acronym disambiguation, where each example consists of a sentence, a short form within that sentence, and the correct long form that the short form refers to, which may or may not appear in the sentence. This dataset was split into 50,034 training examples, 6189 development examples, and 6218 test examples. To construct this dataset, the authors used SciAI to compile a dictionary of acronyms consisting of short forms with their possible long forms. The long forms were normalized through a combination of edit distance and human verification. They then used the one-sense-per-discourse assumption to infer that if a short form is mapped to a long form in the SciAI dataset, then they can find other short forms within the document to use as ambiguous acronyms.

\subsubsection{Duplicates in SciAD}
When exploring the SciAD dataset, we noticed that it contained many duplicates: while there are 62,441 total examples in the dataset, there are only 42,945 examples with a unique (sentence, acronym) pair. For 12,672 of the examples, there exists at least one other example with the exact same sentence and acronym. 45.4\% of the development examples contained a duplicate in the training data, and 45.1\% of the test examples contained a duplicate in the training data. This overlap of data between train time and test time suggests that SciAD is a biased measure of performance on the AD task.

To make matters more interesting, duplicate examples do not always have duplicate labels: 10.5\% of our duplicated examples in the train and development datasets contain more than one long form label. 93.1\% of the development examples that have duplicates within the training dataset share a label with at least one of the duplicates, and 10.8\% of the development examples that have duplicates within the training dataset have a conflicting label with at least one of the duplicates. Since the AD task asks us to find suitable long forms using features extracted from the sentence and short form alone based on the training data, we claim that the accuracy of any model should be upper bounded by 93.1\% on the 45.4\% of the development data that contains a duplicate in the training data.

It is plausible that some of these label conflicts among duplicates is genuine: two different papers could write the exact same sentence yet refer to different acronym expansions. But we suspect that human error from the annotators is the more likely explanation for most of these cases.

In order to remedy this problem, we propose that when one measures development and test performance, they ignore examples that also exist in the training data. In the experiments section, we report our model performance on both this subset of the data as well as the full dataset. Additionally, we propose removing training examples that are duplicates of other training examples for model training. In order to resolve conflicting labels, one can use the more common label among the duplicates. For convenience to other researchers, we released our deduplicated version of the SciAD dataset which we call SciAD-dedupe.\footnote{https://github.com/PrimerAI/sdu-data}

\subsection{Auxiliary AD Data}
To build on the training data provided by SciAD, we took a distantly supervised approach to build a more noisy dataset we call AuxAD. We queried arXiv abstracts for acronyms found in the SciAD dictionary, and collected 56,874 such abstracts. We then assumed that if a short and long form from the dictionary both appeared within a sentence, then the short form can be resolved to the long form. Using the one-sense-per-discourse assumption, we found other sentences within the document that contained the short form and assumed that these short forms also corresponded to the same long form. This resulted in a dataset of 112,788 examples, which we release in our same data repository.

While this AuxAD dataset contains more examples than SciAD, it is less diverse: both datasets started with the same dictionary of 2308 terms, but the SciAD training dataset contains 2152 unique terms and the AuxAD dataset contains 1268 unique terms. We suspect that this is because SciAD used full arXiv documents while AuxAD relied on arXiv abstracts, and thus had a harder time finding certain terms from the dictionary.

\section{Methods}
\subsection{Acronym Identification}
Our model architecture consisted of using a transformer-based language model \cite{transformer} to embed the input sentence tokens, followed by a linear projection onto logits for each BIO tag, an approach mirroring our model for Named Entity Recognition \cite{ner}. We started by joining together the word-tokenized input sentence, and re-tokenizing the sentence with SentencePiece byte-pair encoding \cite{kudo-richardson-2018-sentencepiece} to get $L$ tokens. These $L$ tokens are embedded with the XLNet language model \cite{Yang2019XLNetGA} to get an $\mathbb{R}^H$ embedding for each token, where $H$ is 768 for XLNet-base and 1024 for XLNet-large. We run these embeddings through a linear layer to get $T$ tag logits per token, where $T$ is the number of BIO tags (in this case 5). We use the tag with the highest logit per token as the predicted tag, with the label of the first byte-pair encoded token within a word being used for the word.

\paragraph*{Model Training} Training was performed on both the XLNet encoder and the linear projection weight matrix with cross-entropy loss on output logits and an AdamW optimizer. Training hyperparameters included:
\begin{itemize}
    \item Pretrain on AuxAI then finetune on SciAI, or just train on SciAI
    \item The subset of AuxAI to use when finetuning
    \item XLNet model size
    \item Whether or not to down-weigh the $O$ tag
    \item Learning rate
\end{itemize}
We formed an ensemble of these XLNet models trained with different hyperparameter configurations, and averaged together their predicted logits during inference time. After picking the highest scoring tag per token, we cleaned up the predictions such that $I$ tags could not follow $O$ tags to get our final predictions.

\subsection{Acronym Disambiguation}
\citeauthor{veyseh-et-al-2020-what} modeled AD as a classification problem: given a sentence and a short-form acronym within that sentence, they used a classifier to predict the acronym's expansion. We instead view it as an information retrieval problem: given a test sentence containing an acronym, we want to find the most similar training sentence and use its label. The intuition behind this approach is that contextual clues within a sentence can determine the subfield of research that the paper falls into. By computing the similarity between two sentences, we could perhaps identify if they are within the same research field based on how much their semantics align. For instance, a sentence talking about ``CNN'' would likely include either several machine learning terms or several news terms. We can compare our sentence to several others in the training dataset, and if the dataset is sufficiently comprehensive, we should be able to find a sentence semantically similar to the sentence in question.

More specifically, we start by computing a sentence embedding for every example in the datasets. To infer a label for a given test example, we compute the cosine similarity between its embedding and the embedding of every sentence in the training dataset, pick the training sentence with highest cosine similarity, and use its label. We were able to squeeze out a small performance boost by additionally checking to see if any possible expansion for the acronym appears within the sentence itself, and using that expansion if we find it.

Measuring the utility of this approach is complicated by the fact that the dataset contains many duplicate sentences across testing and training datasets, and sometimes their labels conflict. In cases where multiple duplicate sentences were found, we used the label that was more common overall in the training dataset.

\paragraph*{Model Training}
To train an embedding model for this task, we constructed datasets of sentence pairs from SciAD and AuxAD, where the sentence pairs share a short form acronym and are labeled as having the same long form or a different long form, with a balanced number of positive and negative pairs. We used various transformer-based language models as encoders, and trained these language models as Twin Networks (also known as Siamese Networks) \cite{Chicco2021}: sentence embeddings $(e_1, e_2)$ were computed for sentence pair $(s_1, s_2)$ by running each sentence individually through the same encoder. The cosine similarity between the sentences was computed as
\[
\cos(e_1, e_2) = \frac{e_1^T e_2}{||e_1|| \cdot ||e_2||}
\]

The encoder weights were optimized through mean squared error loss for the sentence pairs representing training examples:
\[
\mathcal{L}_{MSE}(\mathcal{D}) = \frac{1}{n} \sum_{(s_1, s_2, y) \in \mathcal{D}} (y - \cos(E(s_1), E(s_2)))^2
\]
where $\mathcal{D}$ is our dataset of $n$ training examples, $E$ is our transformer embedding model, and $y$ is our desired similarity score, which was 1 if the sentences shared a long form and 0 otherwise. Training our encoder in this way teaches it to learn an embedding space for which sentences containing acronyms with the same meaning will have higher cosine similarity than sentences containing acronyms with different meanings.

\paragraph*{Pretrained Models}
In addition to our trained model, we also tested the other embedding methods of SIF \cite{arora2017asimple} and several pretrained models from sentence transformers \cite{reimers-2019-sentence-bert}:

\begin{itemize}
\item XLM \cite{lample2019cross} trained for paraphrase detection
\item DistilRoBERTa \cite{sanh2020distilbert} trained for paraphrase detection
\item DistilRoBERTa trained for information retrieval on the MS MARCO dataset \cite{bajaj2016ms}
\item DistilRoBERTa trained for Quora question similarity
\end{itemize}

Our final system was an ensemble of these models plus some trained models, where cosine similarity scores were averaged across models in the ensemble.

\section{Acronym Identification Experiments}
\subsection{Model Building}
Our model was implemented using our existing codebase for Named Entity Recognition, which was based on PyTorch Transformers \cite{wolf-etal-2020-transformers}. Each XLNet model took between 10 and 60 minutes to train on a single NVIDIA V100 GPU, depending on hyperparameters like the number of epochs and training dataset. Inference took 3ms per example when using a batch size of 16.

\subsection{Performance}
\begin{table}[ht]
\centering
\begin{tabular}{@{}lccc@{}}
\toprule
\textbf{System} & \textbf{F1} & \textbf{Precision} & \textbf{Recall} \\ \midrule
Baseline & 85.46 & 93.22 & 78.90 \\
LSTM-CRF* & 86.55 & 86.96 & 86.16 \\ \midrule
XLNet, SciAI & 92.17 & 91.62 & 92.72 \\
XLNet, AuxAI & 66.96 & 86.10 & 54.78 \\
XLNet, AuxAI $\rightarrow$ SciAI & 93.14 & 93.23 & 93.04 \\
XLNet Ensemble & \textbf{93.63} & \textbf{93.99} & \textbf{93.28} \\ \bottomrule
\end{tabular}
\caption{Acronym identification performance of various models on the SciAI dataset. Results for the LSTM-CRF model were taken from \citeauthor{veyseh-et-al-2020-what} which used the test dataset, while the other scores are on the development dataset. Our models are below the line, with ``$\rightarrow$'' denoting finetuning. All performance metrics are macro-averaged between short and long forms.}
\label{tab:ai-results}
\end{table}

Performance results are shown in table \ref{tab:ai-results}. In this table, we compare our methods to a rule based baseline \cite{ai-baseline} and the LSTM-CRF model proposed by \citeauthor{veyseh-et-al-2020-what}. All scores are computed on the development dataset due to fact that test dataset labels are not yet publicly available, except for the LSTM-CRF model where scores are taken from their paper. It is clear that while a model trained on just our AuxAI dataset performs poorly, pretraining on AuxAI then finetuning on SciAI results in a measurable boost in performance. Our final ensemble method consisted of 15 different XLNet models trained with different hyperparameters, and achieved an F1 score of 92.60 on the test set.

\paragraph*{Error Analysis}
We performed a small-scale error analysis by looking at a random sample of 50 mistakes made by the ensemble on the SciAI development dataset. Of those mistakes, 18 were genuine mistakes made by the model, 24 were errors made by the human annotators, and 8 were too ambiguous for us to tell. Both model and human mistakes were most commonly the result of failing to extract an acronym that should have been extracted, representing 39 of the errors: the fact that humans frequently missed acronyms within the data likely led to trained models being overly conservative. 7 of the errors came from a misalignment between the true and predicted boundaries of acronyms, and only 1 error came from incorrectly extracting a non-acronym.

\section{Acronym Disambiguation Experiments}
\subsection{Model Building}
Our embedding models were all based on the sentence-transformers library \cite{reimers-2019-sentence-bert}, with the exception of SIF, for which we used fastText \cite{joulin2016fasttext}. Our final ensemble consisted of the following embedding models:

\begin{itemize}
    \item SIF
    \item XLM paraphrase
    \item DistilRoBERTa paraphrase
    \item DistilRoBERTa MS Marco
    \item DistilBERT Quora
    \item RoBERTa SciAD
    \item XLM paraphrase finetuned on AuxAD
\end{itemize}

The last two of these methods were models that we trained as Twin Networks. Training our Twin Network transformer models took around 20 minutes on an NVIDIA V100 GPU depending on what data was used and the number of training epochs. Evaluation consisted of embedding all of the sentences in the training and testing data, which took around 2 minutes per model on a V100, and computing distances between training and testing data, which took around a minute on 16 CPUs for the whole ensemble.

When predicting labels for the development dataset, we used the SciAD training dataset for finding matches, and when predicting labels for the test dataset, we merged together the training and development datasets from SciAD. We experimented with using the AuxAD as well as the SciAD datasets at query time, but found that this led to a slight decrease in performance. Only 12\% of SciAD-dedupe development examples had a closer match in AuxAD than SciAD training, despite AuxAD being a larger dataset, which can largely be explained by the fact that the AuxAD dataset contained fewer terms. Within the small proportion of AuxAD examples that are used, we tend to have less accurate predictions, with an accuracy of 87\% on SciAD-dedupe versus an accuracy of 96\% that we get on the chosen AuxAD examples.

\subsection{Performance}

\begin{table}[ht]
\centering
\begin{tabular}{@{}lcc@{}}
\toprule
\textbf{System} & \textbf{SciAD} & \textbf{SciAD-dedupe} \\ \midrule
Baseline & 59.73 & 59.97 \\
GAD* & 81.90 & - \\ \midrule
SIF & 88.11 & 89.13 \\
XLM paraphrase & 89.42 & 90.89 \\
DistilRoBERTa paraphrase & 89.20 & 90.56 \\
DistilRoBERTa MS Marco & 88.48 & 89.78 \\
DistilBERT Quora & 86.09 & 86.34 \\
RoBERTa SciAD & 88.18 & 89.46 \\
XLM paraphrase $\rightarrow$ AuxAD & 83.75 & 83.04 \\
Ensemble & \textbf{91.22} & \textbf{93.15} \\ \bottomrule
\end{tabular}
\caption{Acronym disambiguation performance of various systems on the SciAD dataset. Results for the GAD model were taken from \citeauthor{veyseh-et-al-2020-what} which used the test dataset, while the other scores were computed for the development dataset. Models below the line are the methods we tested, which represent a combination of pretrained sentence transformers and models we trained ourselves, with ``$\rightarrow$'' denoting finetuning.}
\label{tab:ad-results}
\end{table}

Table \ref{tab:ad-results} shows the macro-averaged F1 scores for each of the individual embedding methods, the embedding ensemble, the GAD classifier proposed in \citeauthor{veyseh-et-al-2020-what}, and the baseline of using the most frequent expansion for an acronym. Performance is shown for the SciAD development set, as well as the development set of SciAD-dedupe. The exception is GAD, for which we include the performance on the test dataset reported in \citeauthor{veyseh-et-al-2020-what}. We can see that the ensemble clearly outperforms the rest of the models, including every individual embedding model it is comprised of. The individual embedding models perform similarly, except for the finetuned XLM paraphrase model. Despite the poor performance from this model, we found that it was valuable to include as a member of the ensemble. Our ensemble achieved an F1 score of 91.58 on the test dataset.

What is also interesting to see is that the systems tend to perform better on the SciAD-dedupe than SciAD, which is counter-intuitive considering the fact that leaking training data into the testing data should theoretically drive up performance scores. To investigate this, we extracted the subset of the SciAD development dataset that was duplicated from the training dataset, and measured the performance of a system that uses the most frequent long form of training examples with the same duplicated tokens. This method achieves an F1 score of 89.41, which is surprisingly lower than the F1 score of our models on the deduplicated data. This subset of the development dataset that repeats sentences from the training dataset thus seems to be quite noisy.

\begin{figure}[ht]
\centering
\includegraphics[width=0.45\textwidth]{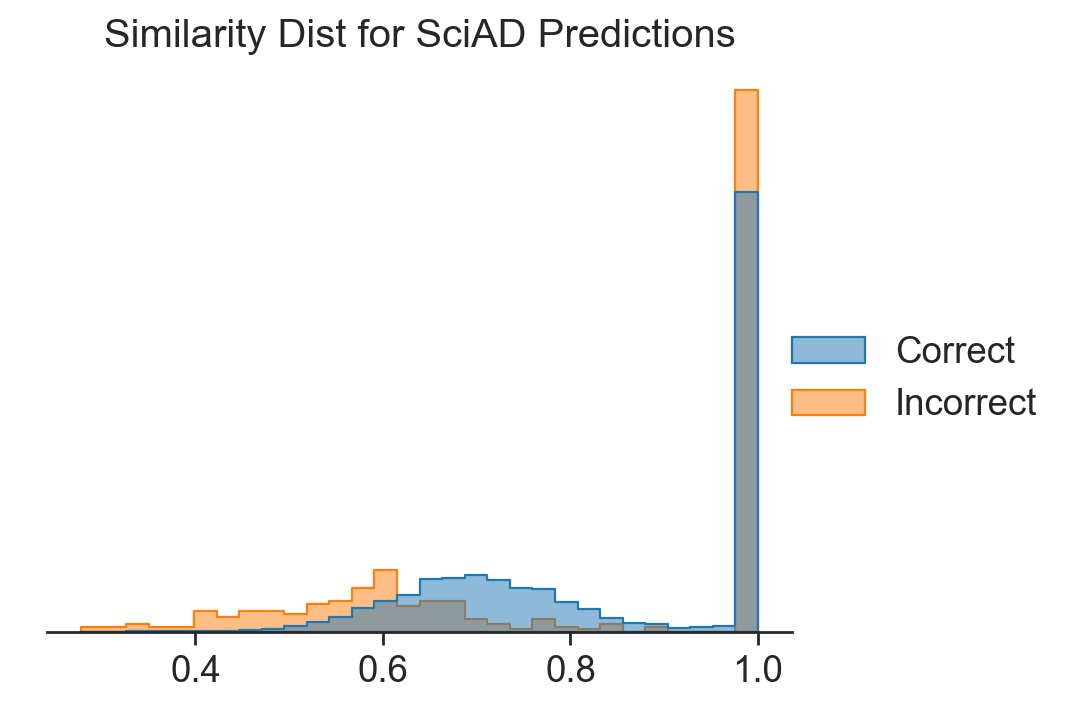}
\caption{Similarity score distributions for predictions on SciAD by our ensemble. Each score represents the cosine similarity between an example in the development dataset and its closest example in the training dataset. The blue distribution is for examples that were judged to be correct, and the orange distribution is for examples that were judged to be incorrect. Both distributions were normalized.}
\label{fig:score-dist}
\end{figure}

Figure \ref{fig:score-dist} shows the distribution of cosine similarity scores between development examples and the testing example inferred to be most similar within the training dataset for our ensemble on SciAD. The distribution of scores for correct examples and incorrect examples are shown and normalized separately. We can see visually that $p(M|\neg C) > p(M|C)$ where $M$ indicates that our development example found a perfect match in the training dataset and $C$ indicates that our predicted expansion is correct. We can also see that if we ignore the perfect matches, correct predictions tend to have higher similarity scores than incorrect predictions, suggesting that our model can trade off recall for boosts in precision by using the similarity scores as a threshold.

\paragraph*{Error Analysis}
We performed a small-scale error analysis by looking at a random sample of 50 mistakes made by the ensemble on the SciAD development dataset. Of those mistakes, 18 were genuine mistakes made by the model, and 32 were mistakes made by human annotators. 
10 of the model mistakes came from semantically similar sentences across development and training having different labels, which highlights a limitation of this approach. 8 of the model mistakes were from the lack of similar training examples for a given development example, which could be potentially fixed given a larger training corpus. 30 of the human mistakes came from conflicting labels in duplicate examples, and the remaining 2 of the human errors came from mistakes during canonicalization of long forms, such as ``sum capacity'' and ``sum capacities'' both existing in the acronym dictionary.

\section{Conclusion}
In conclusion, we have developed new neural models for acronym identification and disambiguation. Our acronym identification model uses a transformer followed by linear projection, and our acronym disambiguation model finds similar examples with embeddings learned from Twin Networks. Both models benefited from ensembling, and both models achieve significant performance gains over the models originally proposed by \citeauthor{veyseh-et-al-2020-what} 

We introduced new datasets for acronym identification and disambiguation, AuxAI and AuxAD, which were labeled through distant supervision. We also identified a duplication issue in the SciAD dataset, and formed a deduplicated version of this dataset that we call SciAD-dedupe. We released all three of these datasets and we hope that they serve as useful tools for the NLP community.

\bibliography{main.bib}
\bibliographystyle{aaai}
\end{document}